\documentclass[runningheads]{llncs}
\usepackage{graphicx,xspace,amsmath,booktabs,multicol,caption,subcaption,wrapfig,hyperref,bbding}
\hypersetup{hidelinks,backref=true,pagebackref=true,hyperindex=true,colorlinks=true,breaklinks=true,urlcolor= blue,bookmarks=true,bookmarksopen=false,pdftitle={`Title},pdfauthor={Author}}
\pdfoutput=1
\newcommand{\ModelName}{{UTRNet}\xspace}
\newcommand{\DatasetNameReal}{{UTRSet-Real}\xspace}
\newcommand{\DatasetNameSynth}{{UTRSet-Synth}\xspace}
\newcommand{\DetectionDataset}{{UrduDoc}\xspace}

\newcommand{\mypara}[1]{\vspace{0.5em} \noindent \textbf{#1:}}

\begin{document}

\title{\ModelName: High-Resolution Urdu Text Recognition In Printed Documents}
\titlerunning{\ModelName: Urdu Text Recognition In Printed Documents}

\author{Abdur Rahman (\Envelope)\orcidID{0000-0002-9547-2435} \and
Arjun Ghosh \and
Chetan Arora}

\authorrunning{A. Rahman et al.}

\institute{Indian Institute of Technology Delhi\\
\email{ch7190150@iitd.ac.in}}

\maketitle 

\let\thefootnote\relax\footnotetext{This is a pre-print of our paper accepted for presentation at \href{https://www.icdar2023.org/}{ICDAR 2023}. It has not undergone peer review or any post-submission improvements. The Version of the Record of this contribution is published in “Document Analysis and Recognition - ICDAR 2023” and is available online at \url{https://doi.org/10.1007/978-3-031-41734-4_19}. We kindly request that readers refer to the published version of the record of our work for the most accurate and authoritative representation.}

\begin{abstract}
	In this paper, we propose a novel approach to address the challenges of printed Urdu text recognition using high-resolution, multi-scale semantic feature extraction. Our proposed \ModelName architecture, a hybrid CNN-RNN model, demonstrates state-of-the-art performance on benchmark datasets. To address the limitations of previous works, which struggle to generalize to the intricacies of the Urdu script and the lack of sufficient annotated real-world data, we have introduced the \DatasetNameReal, a large-scale annotated real-world dataset comprising over 11,000 lines and \DatasetNameSynth, a synthetic dataset with 20,000 lines closely resembling real-world and made corrections to the ground truth of the existing IIITH dataset, making it a more reliable resource for future research. We also provide {\DetectionDataset}, a benchmark dataset for Urdu text line detection in scanned documents. Additionally, we have developed an online tool for end-to-end Urdu OCR from printed documents by integrating \ModelName with a text detection model. Our work not only addresses the current limitations of Urdu OCR but also paves the way for future research in this area and facilitates the continued advancement of Urdu OCR technology. The project page with source code, datasets, annotations, trained models, and online tool is available at \href{https://abdur75648.github.io/UTRNet/}{abdur75648.github.io/UTRNet}.
	
\end{abstract}

\section{Introduction}
\label{section:intro_section}

Printed text recognition, also known as optical character recognition (OCR), involves converting digital images of text into machine-readable text and is an important topic of research in the realm of document analysis with applications in a wide variety of areas \cite{ocrapplications1}. While OCR has transformed the accessibility \& utility of written/printed information, it has traditionally been focused on Latin languages, leaving non-Latin low-resource languages such as Urdu, Arabic, Pashto, Sindhi and Persian largely untapped. Despite recent developments in Arabic script OCR \cite{surveyArabicOCR,7926062,surveyArabicOCR_2023}, research on OCR for Urdu remains limited \cite{surveyurdu1,surveyurdu2,surveyurdu3}. With over 230 million native speakers and a huge literature corpus, including classical prose and poetry, newspapers, and manuscripts, Urdu is the 10th most spoken language in the world \cite{Mushtaq2021-cn,iiith17urdu}. Hence the development of a robust OCR system for Urdu remains an open research problem and a crucial requirement for efficient storage, indexing, and consumption of its vast heritage, mainly its classical literature.

\begin{figure}[t]
	\begin{minipage}[b]{\textwidth}
		\centering
		\centerline{\includegraphics[width=\textwidth]{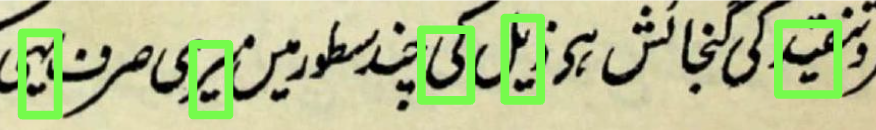}}
		\centerline{(a) Sample 1}\medskip
	\end{minipage}
	\begin{minipage}[b]{.35\textwidth}
		\centering
		\centerline{\includegraphics[width=\linewidth]{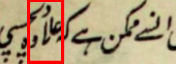}}
		\centerline{(b) Sample 2}\medskip
	\end{minipage}
	\hfill
	\begin{minipage}[b]{.3\textwidth}
		\centering
		\centerline{\includegraphics[width=\linewidth]{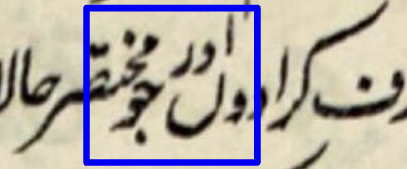}}
		\centerline{(c) Sample 3}\medskip
	\end{minipage}
	\hfill
	\begin{minipage}[b]{.325\textwidth}
		\centering
		\centerline{\includegraphics[width=\linewidth]{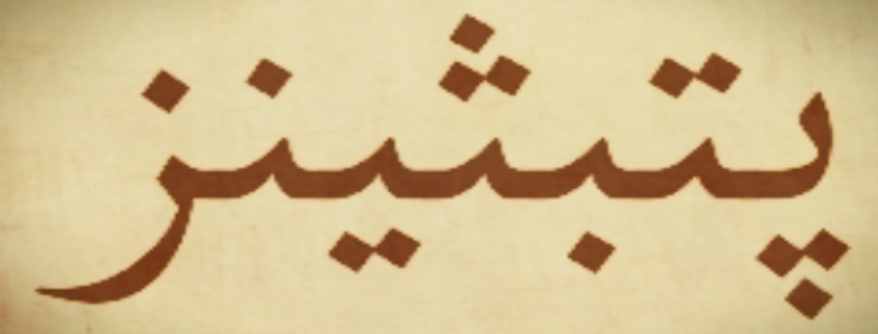}}
		\centerline{(d) Sample 4}\medskip
	\end{minipage}
	\caption{Intricacies of the script. (a) All 5 characters inside the boxes are the same, but they look different as the shape of the character relies on its position and context in the ligature. The box contains 8 characters in (b) and 11 in (c), demonstrating that the script has a very high degree of overlap. (c) The word shown consists of 6 distinct characters of similar shape, which differ merely by the arrangement of little dots (called \textit{``Nuqta''}) around them.}
	\label{fig:intricacies}
\end{figure}

\begin{figure}[t]
	\includegraphics[width=\linewidth]{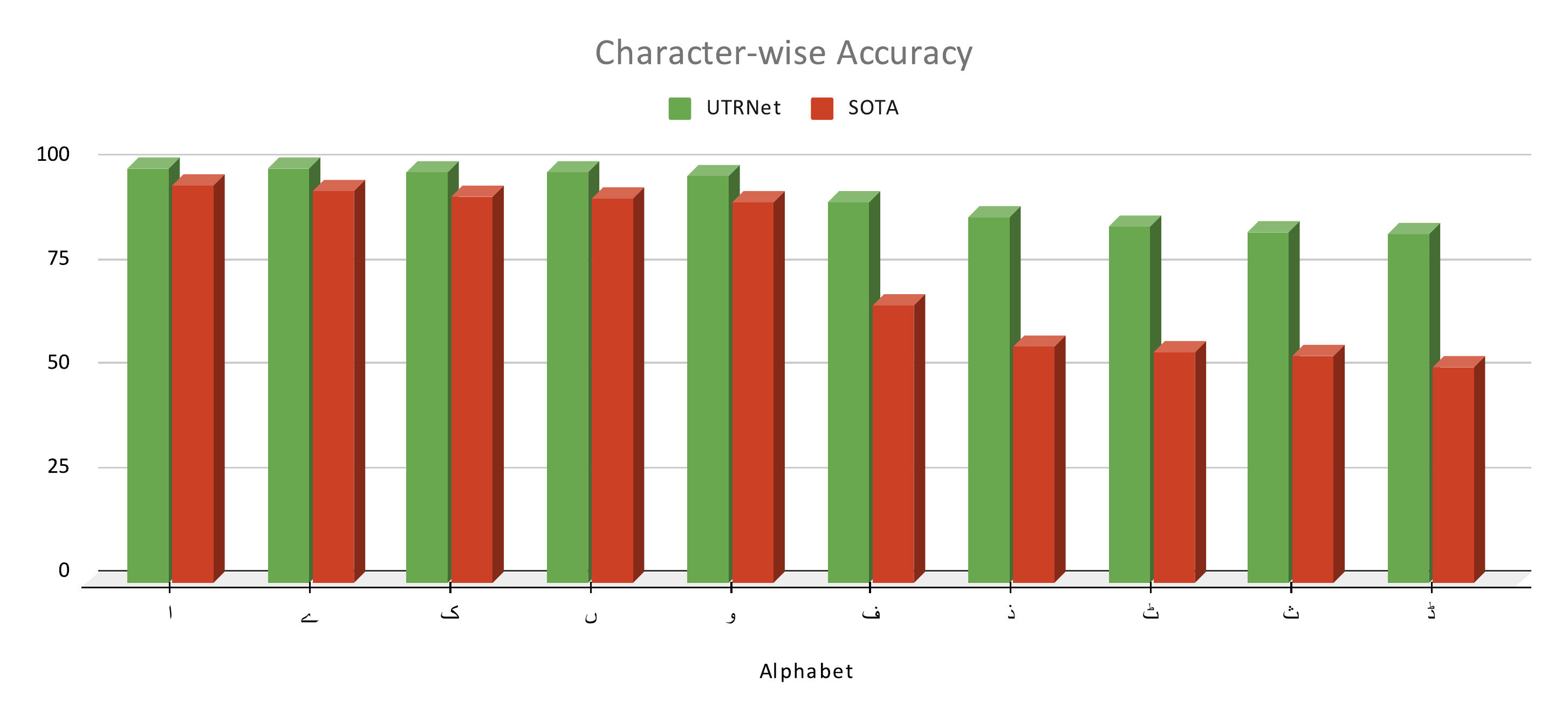}
	\caption{Plot showing character-wise accuracies of \ModelName-Small and SOTA for Urdu OCR, presented in \cite{iiith17urdu}. It can be observed that the accuracy gap is larger for the last 5 characters (right side), which differ from several other characters having the same shape only in terms of presence and arrangement of dots around them, as compared to the first 5 characters, which are simpler ones.
	}
	\label{fig:char_accuracy}
\end{figure}

However, the intricacies of the Urdu script, which predominantly exists in the Nastaleeq style, present significant challenges. It is primarily cursive, with a wide range of variations in writing style and a high degree of overall complexity, as shown in Fig. \ref{fig:intricacies}. Though Arabic script is similar \cite{arabic_intricacies}, the challenges faced in recognizing Urdu text differ significantly. Arabic text is usually printed in the Naskh style, which is mostly upright and less cursive, and has only 28 alphabets \cite{surveyArabicOCR_2023}, in contrast to the Urdu script, which consists of 45 main alphabets, 26 punctuation marks, 8 honorific marks, and 10 Urdu digits, as well as various special characters from Persian and Arabic, English alphabets, numerals, and punctuation marks, resulting in a total of 181 distinct glyphs \cite{iiith17urdu} that need to be recognized. Furthermore, the lack of large annotated real-world datasets in Urdu compounds these challenges, making it difficult to compare different models' performance accurately and to continue advancing research in the field (Section \ref{section:dataset_section}). The lack of standardization in many Urdu fonts and their rendering schemes (particularly in early Urdu literature) further complicates the generation of synthetic data that closely resembles real-world representations. This hinders experiments with more recent transformer-based OCR models that require large training datasets (Table \ref{tab:sota-comparisions}B) \cite{transformerurduocr,tttvit,vitpaper}. As a result, a naive application of the methods developed for other languages does not result in a satisfactory performance for Urdu (Table \ref{tab:sota-comparisions}), highlighting the need for exclusive research in OCR for Urdu.

The purpose of our research is to address these long-standing limitations in printed Urdu text recognition through the following key contributions:
\begin{enumerate}
	\item We propose a novel approach using high-resolution, multi-scale semantic feature extraction in our \ModelName architecture, a hybrid CNN-RNN model, that demonstrates state-of-the-art performance on benchmark datasets.
	\item We create \DatasetNameReal, a large-scale annotated real-world dataset comprising over 11,000 lines.
	\item We have developed a robust synthetic data generation module and release \DatasetNameSynth, a high-quality synthetic dataset of 20,000 lines closely resembling real-world representations of Urdu text.
	\item We correct many annotation errors in one of the benchmark datasets \cite{iiith17urdu} for Urdu OCR, thereby elevating its reliability as a valuable resource for future research endeavours and release the corrected annotations publicly.
	\item We have curated {\DetectionDataset}, a real-world Urdu documents text line detection dataset. The dataset is a byproduct of our efforts towards \DatasetNameReal, and contains line segmentation annotation for 478 pages generated from more than 130 books. 
	\item To make the output of our project available to a larger non-computing research community, as well as lay users, we have developed an online tool for end-to-end Urdu OCR, integrating \ModelName with a third-party text detection model.
\end{enumerate}

In addition to our key contributions outlined above, we conduct a thorough comparative analysis of state-of-the-art (SOTA) Urdu OCR models under similar experimental setups, using a unifying framework introduced by \cite{clovaai} that encompasses feature extraction, sequence modelling and prediction stages and provides a common perspective for all the existing methods. We also examine the contributions of individual modules towards accuracy as an ablation study (Table \ref{tab:ablation}). Finally, we discuss the remaining limitations and potential avenues for future research in Urdu text recognition.

\section{Related Work}
\label{section:related_work_section}

The study of Urdu OCR has gained attention only in recent years. While the first OCR tools were developed more than five decades back \cite{surveyurdu1}, the earliest machine learning based Urdu OCR was developed in 2003 \cite{firsturduocr}. Since then, the research in this field has progressed from isolated character recognition to word/ligature level recognition to line level recognition (see \cite{surveyurdu1,surveyurdu2,surveyurdu3} for a detailed survey). Early approaches primarily relied on handcrafted features, and traditional machine learning techniques, such as nearest neighbour classification, PCA \& HMMs \cite{char_urdu_ocr2,char_urdu_ocr3,old_urdu0cr_1,lm_urdu_ocr}, to classify characters after first segmenting individual characters/glyphs from a line image. These techniques often required extensive pre-processing and struggled to achieve satisfactory performance on large, varied datasets. 

Recently segmentation-free end-to-end approaches based on CNN-RNN hybrid networks \cite{iiith17urdu} have been introduced, in which a CNN \cite{cnn_for_features} is used to extract low-level visual features from input data which is then fed to an RNN \cite{rnn_for_sequence} to get contextual features for the output transcription layer. Among the current SOTA models for Urdu OCR (Table \ref{tab:sota-comparisions}C), VGG networks \cite{vgg_orig} have been used for feature extraction in \cite{iiith17urdu,mdlstm_urdu_ocr,mdlstm_urdu_ocr2}, whereas ResNet networks \cite{resnet_orig} have been utilized in \cite{resnet_urdu_ocr}. For sequential modelling, BiLSTM networks \cite{bilsm_orig} have been used in \cite{iiith17urdu,resnet_urdu_ocr}, while MDLSTM networks \cite{mdlstm_orig} have been employed in \cite{mdlstm_urdu_ocr,mdlstm_urdu_ocr2}. All of these approaches utilize a Connectionist Temporal Classification (CTC) layer \cite{ctc_paper} for output transcription. In contrast, \cite{gru_attn_paper} utilizes a DenseNet \cite{densenet_orig} and GRU network \cite{gru_orig} with an attention-based decoder layer \cite{attn_decoder_orig} for final transcription. Arabic, like Urdu, shares many similarities in the script as discussed above, and as such, the journey of OCR development has been similar \cite{iiith17arabic,surveyArabicOCR}. Recent works have shown promising results in recognising Arabic text through a variety of methods, including traditional approaches \cite{isolated_arabic_2017,seg_hybrid_arabic_2017}, as well as DL-based approaches such as CNN-RNN hybrids \cite{arabic_hybrid_cnn_rnn_blstm,iiith17arabic}, attention-based models \cite{arabic_attention_encoder_decoder_2021,Arabic_attention_cnn_rnn}, and a range of network architectures \cite{arabic_mdlstm_lr,arabic_mdlstm_2008,arabic_mdbsltm_2018,arabic_OCFormer}. However, these approaches still struggle when it comes to recognizing Urdu text, as evident from the low accuracies achieved by SOTA Arabic OCR methods like \cite{arabic_hybrid_cnn_rnn_blstm}, \cite{arabic_attention_encoder_decoder_2021} and \cite{Arabic_attention_cnn_rnn} in our experimental results presented in Table \ref{tab:sota-comparisions}.

While each of the methods proposed so far has claimed to have pushed the boundary of the technology, they often rely on the same methodologies used for other languages without considering the complexities specific to Urdu script and, as such, do not fully utilize the potential in this field. Additionally, a fair comparison among these approaches has been largely missing due to inconsistencies in the datasets used, as described in the dataset section below.

\section{Proposed Architecture}
\label{section:architecture_section}

\begin{figure}[t]
	\begin{subfigure}[b]{1.0\linewidth}
		\centering
		\includegraphics[width=1.0\linewidth]{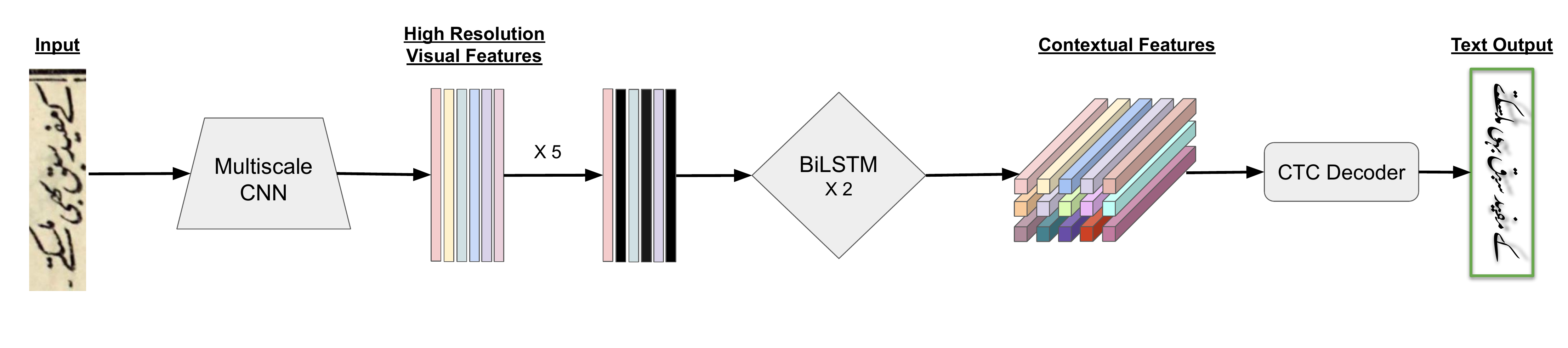}
		\caption{Proposed overall architecture}
		\label{fig:arch_overall}
	\end{subfigure}
	\begin{subfigure}[b]{1.0\linewidth}
		\centering
		\includegraphics[width=1.0\linewidth]{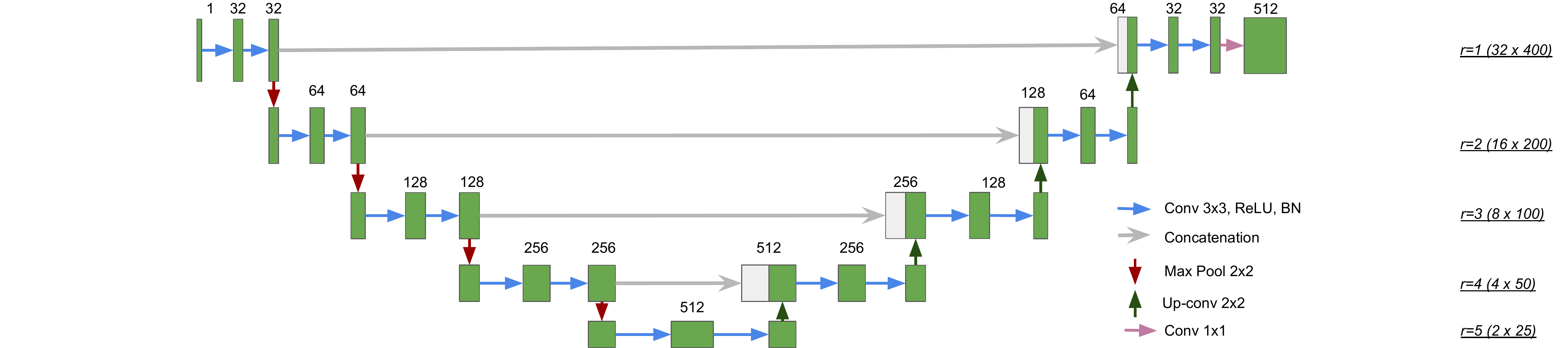}
		\caption{Multiscale feature extraction module for \ModelName-Small based on UNet \cite{unet_orig_paper}}
		\label{fig:arch_small}
	\end{subfigure}
	\begin{subfigure}[b]{1.0\linewidth}
		\centering
		\includegraphics[width=1.0\linewidth]{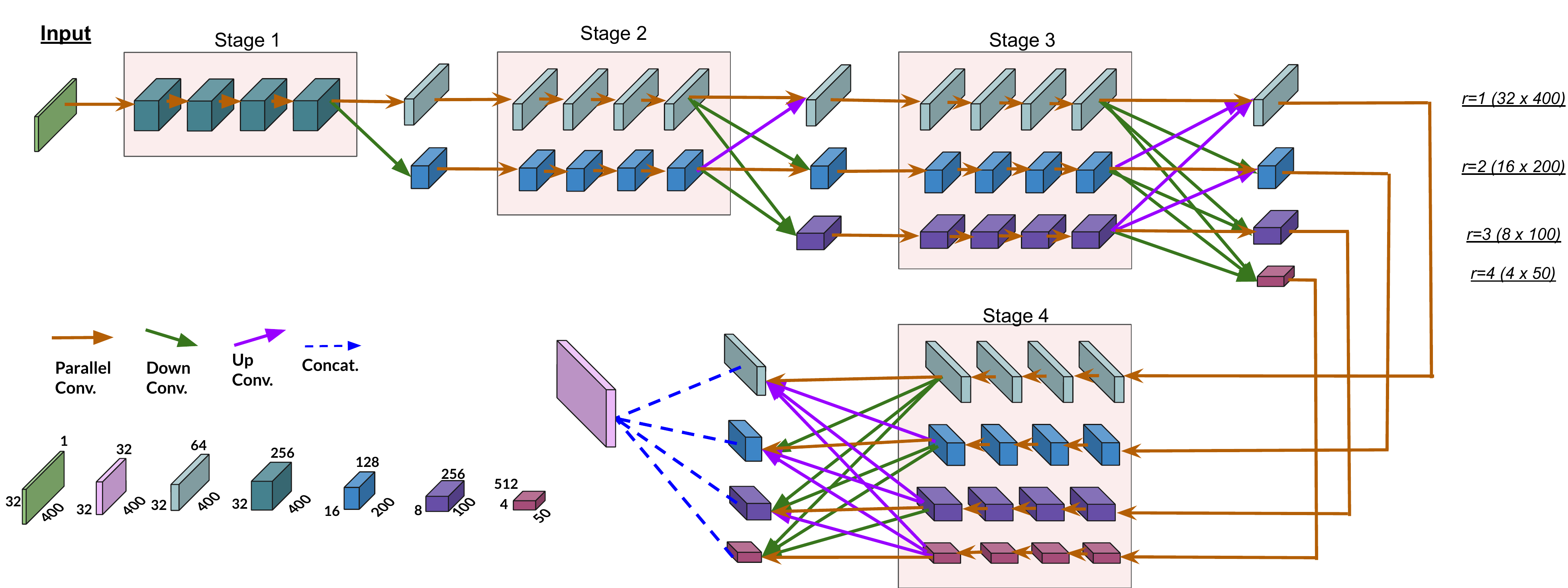}
		\caption{Multiscale feature extraction module for \ModelName-Large based on HRNet \cite{hrnet_orig}}
		\label{fig:arch_large}
	\end{subfigure}
	\caption{Proposed Architecture}
	\label{fig:arch}
\end{figure}

Recently, transformer-based models have achieved state-of-the-art performance on various benchmarks \cite{vit_applications_survey}. However, these models have the drawback of being highly data-intensive and requiring large amounts of training data, making it difficult to use them for several tasks with limited real-world data, such as printed Urdu text recognition (as discussed in Section \ref{section:intro_section}). In light of this, we propose \ModelName (Figure \ref{fig:arch}), a novel CNN-RNN hybrid network that offers a viable alternative. The proposed model effectively extracts high-resolution multi-scale features while capturing sequential dependencies, making it an ideal fit for Urdu text recognition in real-world scenarios. We have designed \ModelName in two versions: \ModelName-Small (10.7M parameters) and \ModelName-Large (47.3M parameters). The architecture consists of three main stages: feature extraction, sequential modelling, and decoding. The feature extraction stage makes use of convolutional layers to extract feature representations from the input image. These representations are then passed to the sequential modelling stage to learn sequential dependencies among them. Finally, the decoding module converts the sequential data feature thus obtained into the final output sequence.

\subsection{High-Resolution Multiscale Feature Extraction}
\label{architecture-high-resolution-cnn}

In our proposed method, we address the wide mismatch in the accuracy of different characters observed in existing techniques, as shown in Figure \ref{fig:char_accuracy}. We posit that this is due to the lack of attention given to small features associated with most Urdu characters by the existing methods. These methods rely upon using standard CNNs, such as RCNN\cite{rcnn_orig}, VGG\cite{vgg_orig}, ResNet\cite{resnet_orig}, etc., as their backbone. However, a significant drawback of using these CNNs is the low resolution representation generated at the output layer. The representation lacks low-level feature information, such as the dots (called \textit{``Nuqta''}) in Urdu characters. To overcome this limitation, in our proposed method, we propose to use a high-resolution multiscale feature extraction technique to extract features from an input image while preserving the small details of the image.

\mypara{\ModelName-Small}
To address the issue of computational efficiency, we propose a lighter variant of our novel \ModelName architecture, referred to as \ModelName-Small, which employs a standard U-Net model (as shown in Figure \ref{fig:arch_small}), initially proposed in \cite{unet_orig_paper} for biomedical image segmentation. The lighter version proposed by us addresses captures high resolution feature maps using the standard U-Net model, originally proposed in \cite{unet_orig_paper} for biomedical image segmentation. We first encode the low-resolution representation of input image \(X\), which captured context from a large receptive field. We then recovers the high-resolution representation using learnable convolutions from the previous decoder layer, and skip connections from the encoder layers. For any resolution with index $r \in \{1,2,3,4,5\}$, the feature map at that resolution can be defined as: $F_r = \text{downsample}(F_{r-1})$ during downsampling. Here $\text{downsample}(F_{r-1})$ is the downsampled feature map from the resolution index $(r-1)$. Similarly, $F_r = \text{concat}(\text{upsample}(F_{r+1}), M_{r})$ represents feature map during upsampling, where $\text{upsample}(F_{r+1})$ is the upsampled feature map from the resolution index $(r+1)$, and $M_{r}$ is the feature map from the downsampling path corresponding to that resolution). This allows the model to aggregate image features from multiple image scales.

\mypara{\ModelName-Large}
The model (Figure \ref{fig:arch_large}) maintains high-resolution representation throughout the process and captures the fine-grained details more efficiently, using an HRNet architecture \cite{hrnet_orig}. The resulting network consists of 4 stages, with the $I^\text{th}$ stage containing streams coming from $\#I$ different resolutions and giving out $\#I$ streams corresponding to the different resolutions. Each stream in the $I^\text{th}$ stage is represented as $R^{I}_{r}$, where $r \in \{1,2,\ldots,I\}$. These streams which are then inter-fused among themselves to get $\#(I+1)$ final output streams, $R^{I,O}_{r}$ for the next stage:
\[
R^{I,O}_{r}= \sum_{i=1}^{I} f_{ir}(R^{I}_{i}) , \qquad r \in \{1,2,\ldots, I+1\} 
\]
The transform function $f_{ir}$ is dependent on the input resolution index $i$ and the output resolution index $r$. If $x=r$, then $f_{xr}(R)=R$. However, if $x<r$, then $f_{xr}(R)$ downsamples the input representation $R$. Similarly, if $x>r$, then $f_{xr}(R)$ upsamples the input resolution. \ModelName-Large uses repeated multi-resolution fusions which allows effective exchange of information across multiple resolutions. Thus, giving a multi-dimensional and high-resolution feature representation of the input image, which is semantically richer.

\subsection{Sequential Modeling And Prediction}

The output from the feature extraction stage is a feature map $V = \{v_i\}$. To prevent over-fitting, we implement a technique called Temporal Dropout \cite{temporal_orig}, in which we randomly drop half of the visual features before passing them to the next stage. We do this 5 times in parallel and take the average. In order to capture the rich contextual information and temporal relationships between the features thus obtained, we pass it through 2 layers of BiLSTM \cite{bilsm_orig} (DBiLSTM) \cite{crnn_orig}. Each BiLSTM layer identifies two hidden states, $h^{f}_t$ and $h^{b}_t$, calculated forward and backward through time, respectively, which are combined to determine one hidden state $h^{t}$, using an FC layer. The sequence $H=\{h_t\}=\text{DBiLSTM}(V)$) thus obtained has rich contextual information from both directions, which is crucial for Urdu text recognition, especially because the shape of each character depends upon characters around it. The final prediction output $Y = \{y1, y2, \ldots\}$, a variable-length sequence of characters, is generated by the prediction module from the input sequence $H$. For this, we use the Connectionist temporal classification (CTC), as described in \cite{ctc_paper}.

\section{Current Publicly Available and Proposed Datasets}
\label{section:dataset_section}

The availability of datasets for the study of Urdu Optical Character Recognition (OCR) is limited, with only a total of six datasets currently available: UPTI \cite{upti_paper}, IIITH \cite{iiith17urdu}, UNHD \cite{unhd_paper}, CENPARMI \cite{cenparmi_paper}, CALAM \cite{calam_paper}, and PMU-UD \cite{pmu_ud_paper}. Of these datasets, only IIITH and UPTI contain printed text line samples, out of which only UPTI has a sufficient number of samples for training. However, the UPTI dataset is synthetic in nature, with limited diversity and simplicity in comparison to real-world images (See Figure \ref{fig:datasets_samples_and_stats}). There is currently no comprehensive real-world printed Urdu OCR dataset available publicly for researchers. As a result, different studies have used their own proprietary datasets, making it difficult to determine the extent to which proposed models improve upon existing approaches. This lack of standardisation and transparency hinders the ability to compare the performance of different models accurately and to continue advancing research in the field, which our work aims to tackle with the introduction of two new datasets. The following are the datasets used in this paper (also summarized in Figure \ref{fig:datasets_samples_and_stats}):

\begin{figure}[t]
	\begin{subfigure}[b]{0.5\linewidth}
		\centering
		\includegraphics[width=1.0\linewidth]{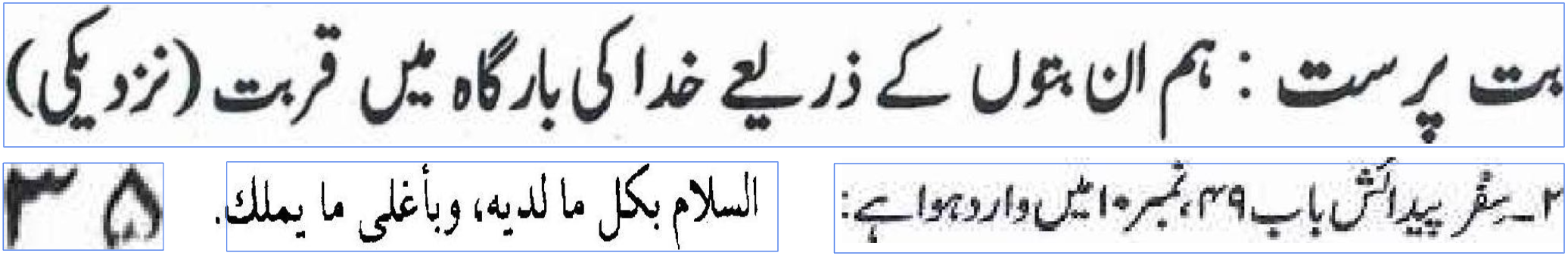}
		\caption{IIITH \cite{iiith17urdu}}
	\end{subfigure}
	\begin{subfigure}[b]{0.5\linewidth}
	\centering
	\includegraphics[width=1.0\linewidth]{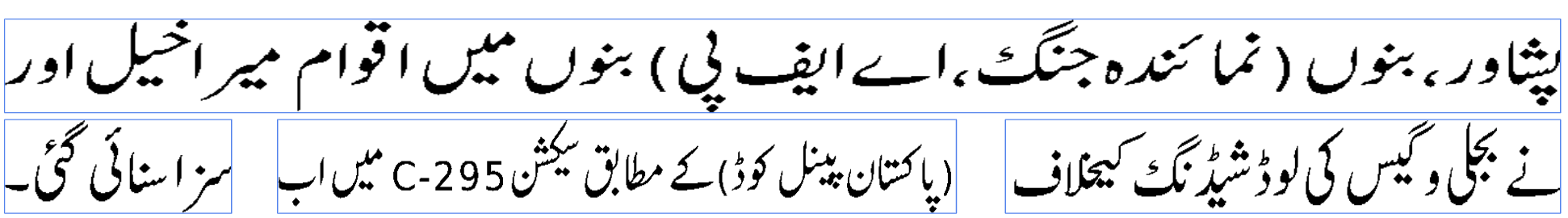}
	\caption{UPTI \cite{upti_paper}}
	\end{subfigure}
	\begin{subfigure}[b]{1.0\linewidth}
	\centering
	\includegraphics[width=1.0\linewidth]{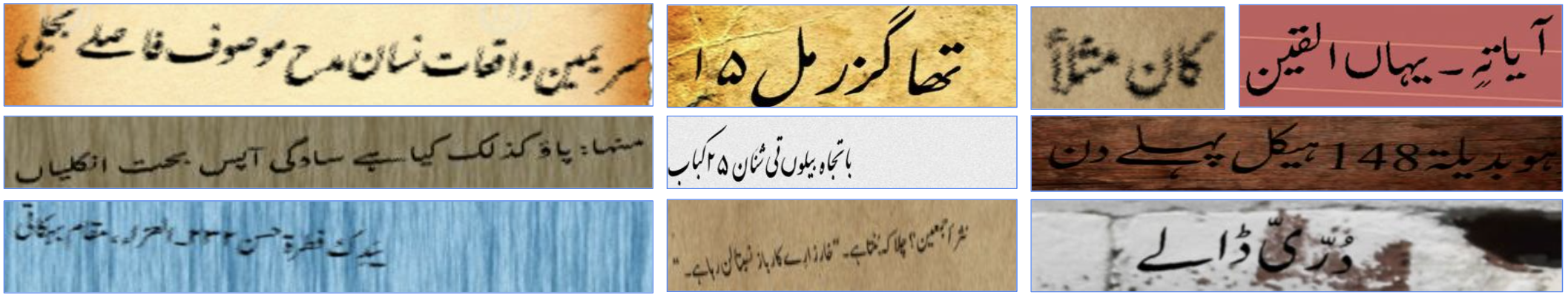}
	\caption{Proposed \DatasetNameSynth}
	\end{subfigure}
	\begin{subfigure}[b]{1.0\linewidth}
	\centering
	\includegraphics[width=1.0\linewidth]{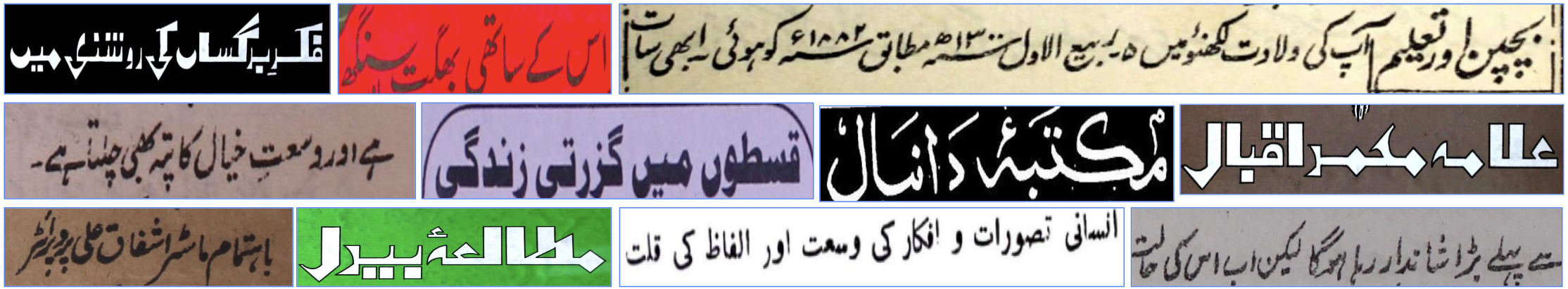}
	\caption{Proposed \DatasetNameReal}
	\end{subfigure}
	\setlength{\tabcolsep}{6pt}
	\begin{tabular}{lcccc}
	\\
	\toprule[1.5pt]
	\textbf{Dataset} & \textbf{Training Set} & \textbf{Validation Set} & \textbf{Vocab Length} & \textbf{Type}\\ [0.5ex]
	\midrule[0.5pt]
	IIITH \cite{iiith17urdu} & NA & 1,610 & 5,772 & Real \\
	UPTI \cite{upti_paper} & 8,051 & 2,012 & 12,054 & Synthetic \\[0.5ex]
	\DatasetNameReal & 9,065 & 2,096 & 22,964 & Real \\
	\DatasetNameSynth & 20,000 & NA & 28,187 & Synthetic \\
	\bottomrule[1.5pt]
	\end{tabular}
	\caption{Sample images, and statistics of the publicly available and proposed datasets. Notice the richness and diversity in the proposed \DatasetNameReal and \DatasetNameSynth datasets, as compared to existing datasets}
	\label{fig:datasets_samples_and_stats}
	\vspace{-1em}
\end{figure}

\begin{figure}[t]
	\includegraphics[width=\linewidth]{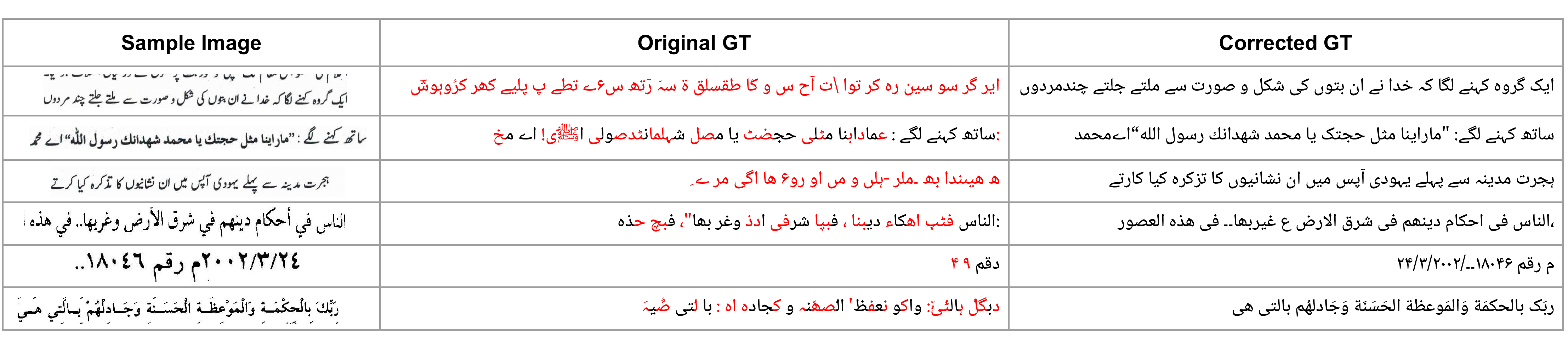}
	\caption{Mistakes in IIITH dataset annotations and our corrections.}
        \label{fig:iiith_mistakes}
\end{figure}

\mypara{IIITH}
This dataset was introduced by \cite{iiith17urdu} in 2017, and it contains only the validation data set of 1610 line images in nearly uniform colour and font. No training data set has been provided. We corrected the ground-truth annotations for this dataset, as we found several mistakes, as highlighted in Figure \ref{fig:iiith_mistakes}.

\mypara{UPTI}
Unlike the other two, this is a synthetic data set introduced in 2013 by \cite{upti_paper}. It consists of a total of 10,063 samples, out of which 2,012 samples are in the validation set. This data set is also uniform in colour and font and has a vocabulary of 12,054 words.

\mypara{Proposed \DatasetNameReal}
A comprehensive real-world annotated dataset curated by us, containing a few easy and mostly hard images (as illustrated in Figure \ref{fig:intricacies}). To create this dataset, we collected 130 books and old documents, scanned over 500 pages, and manually annotated the scanned images with line-wise bounding boxes and corresponding ground truth labels. After cropping the lines and performing data cleaning, we obtained a final dataset of 11,161 lines, with 2,096 lines in the validation set and the remaining in the training set. This dataset stands out for its diversity, with various fonts, text sizes, colours, orientations, lighting conditions, noises, styles, and backgrounds represented in the samples, making it well-suited for real-world Urdu text recognition.

\mypara{Proposed \DatasetNameSynth}
To complement the real-world data in \DatasetNameReal for training purposes, we also present \DatasetNameSynth, a high-quality synthetic dataset of 20,000 lines with over 28,000 total unique words, closely resembling real-world representations of Urdu text. This dataset is generated using a custom-designed synthetic data generation module that allows for precise control over variations in font, text size, colour, resolution, orientation, noise, style, background etc. The module addresses the challenge of standardizing fonts by collecting and incorporating over 130 diverse fonts of Urdu after making corrections to their rendering schemes. It also addresses the limitation of current datasets, which have very few instances of Arabic words and numerals, Urdu digits etc., by incorporating such samples in sufficient numbers. Additionally, it generates text samples by randomly selecting words from a vocabulary of 100,000 words. The generated \DatasetNameSynth has 28,187 unique words with an average word length of 7 characters. The data generation module has been made publicly available on the project page to facilitate further research.

\mypara{Proposed \DetectionDataset}
In addition to the recognition datasets discussed above, we also present {\DetectionDataset}, a benchmark dataset for Urdu text line detection in scanned documents. To the best of our knowledge, this is the first dataset of its kind \cite{urdu_text_det,cursive_text_det}. It was created as a byproduct of the \DatasetNameReal dataset generation process, in which the pages were initially scanned and then annotated with horizontal bounding boxes in COCO format \cite{coco_orig} to crop the text lines. Comprising of 478 diverse images collected from various sources such as books, documents, manuscripts, and newspapers, it is split into 358 pages for training and 120 for validation. The images include a wide range of styles, scales, and lighting conditions, making them a valuable resource for the research community. The {\DetectionDataset} dataset will serve as a valuable resource for the research community, advancing research in Urdu document analysis. We also provide benchmark results of a few SOTA text detection models on this dataset using precision, recall, and h-mean, as shown in Table \ref{tab:detection-results}. The results demonstrate that the ContourNet model \cite{contournet_orig} outperforms the other models in terms of h-mean. It is worth noting that as text detection was not the primary focus of our research but rather a secondary contribution, a thorough examination of text detection has not been conducted. This aspect can be considered a future work for researchers interested in advancing the field of Urdu document analysis. We will make this dataset publicly available to the research community for non-commercial, academic, and research purposes associated with Urdu document analysis, subject to request and upon execution of a no-cost license agreement.

\begin{table}[b]
\centering
\setlength{\tabcolsep}{20pt}
\begin{tabular}{lccc}
	\toprule[1.5pt]
	\textbf{Methods} & \textbf{Precision} & \textbf{Recall} & \textbf{Hmean}\\ [0.5ex]
	\midrule[0.5pt]
	EAST \cite{east_orig} & 70.43 & 72.56 & 71.48 \\
	PSENet \cite{psenet_orign} & 78.32 & 77.91 & 78.11 \\
	DRRG \cite{drrg_orig} & 83.05 & 84.72 & 83.87 \\
	ContourNet \cite{contournet_orig} & 85.36 & 88.68 & 86.99 \\
	\bottomrule[1.5pt] 
\end{tabular}
\vspace{1em}
\caption{Experimental results on {\DetectionDataset}}
\label{tab:detection-results}
\vspace{-2em}
\end{table}

\begin{figure}[t]
	\label{fig:urdudoc_samples}
	\includegraphics[width=\linewidth]{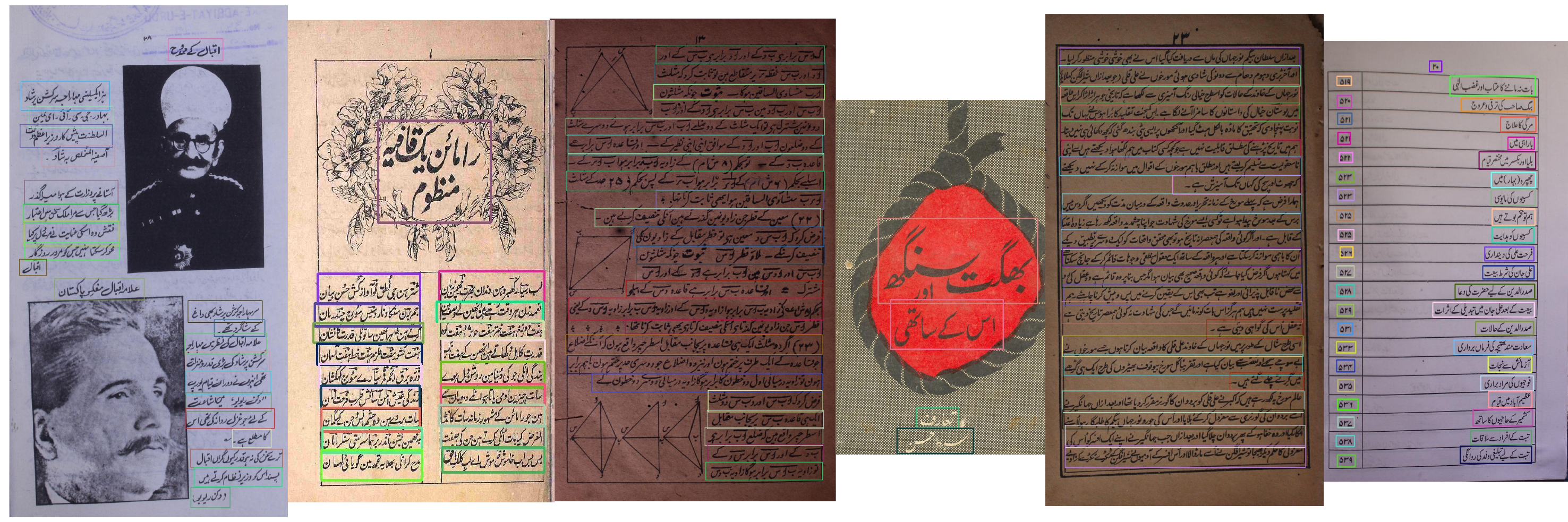}
	\caption{Sample images from the {\DetectionDataset} Dataset: An annotated real-world benchmark for Urdu text line detection in scanned documents}
\end{figure}

\begin{table}[t]
\centering
\setlength{\tabcolsep}{10pt}
\begin{tabular}{lccc}
	\toprule[1.5pt]
	\textbf{Model/Strategy} & \textbf{\DatasetNameReal} & \textbf{IIITH} & \textbf{UPTI} \\ 
	\midrule[1.5pt]
	\multicolumn{4}{c}{Various Multiscale Feature Extraction Backbones} \\
	\midrule[0.5pt]
	UNet \cite{unet_orig_paper} & 90.87 & 86.35 & 95.08 \\
	AttnNet \cite{attn_unet} & 91.95 & 86.45 & 95.26 \\
	ResidualUNet \cite{res_unet} & 91.90 & 87.16 & 95.23 \\
	InceptionUNet \cite{inception_unet} & 92.10 & 87.37 & 95.61 \\
	UNetPlusPlus \cite{unet_plus_plus} & 92.53 & 87.36 & 95.84 \\
	HRNet \cite{hrnet_orig} & \textbf{92.97} & \textbf{88.01} & \textbf{95.97} \\ 
	\midrule[0.5pt]
	\multicolumn{4}{c}{Various Sequential Modelling Backbones} \\
	\midrule[0.5pt]
	LSTM & 91.20 & 87.21 & 94.41 \\
	GRU & 91.48 & 87.04 & 94.58 \\
	MDLSTM & 91.51 & 87.38 & 94.67 \\
	BiLSTM & 91.53 & 87.67 & 94.69 \\
	DBiLSTM & \textbf{92.97} & \textbf{88.01} & \textbf{95.97} \\ 
	\midrule[0.5pt]
	\multicolumn{4}{c}{Various Strategies to Improve Generalization} \\
	\midrule[0.5pt]
	None & 91.07 & 86.80 & 95.07 \\
	Augmentation & 92.35 & 87.53 & 95.48\\
	Augmentation + Temporal Dropout & \textbf{92.97} & \textbf{88.01} & \textbf{95.97} \\  
	\bottomrule[1.5pt] \\
\end{tabular}
\caption{The results of the ablation study for {\ModelName} provide a comprehensive examination of the impact of individual feature extraction and sequence modelling stages and augmentation strategies on the overall performance of the model. By evaluating each stage one-by-one while keeping the remaining stages constant, the results highlight the key factors driving the accuracy of the model, offering valuable insight into optimizing the performance of UTRNet.}
\label{tab:ablation}
\vspace{-2em}
\end{table}

\begin{table}[t]
	\centering
	\setlength{\tabcolsep}{15pt}
	\begin{tabular}{lccc}
		\toprule[1.5pt]
		\textbf{Models} & \textbf{\DatasetNameReal} & \textbf{IIITH} & \textbf{UPTI} \\
		\midrule[1.5pt]
		\multicolumn{4}{c}{\textbf{A.} Baseline OCR models (Hybrid CNN-RNN)} \\
		\midrule[0.5pt]
		R2AM \cite{r2am} & 84.12 & 81.39 & 92.07 \\
		CRNN \cite{crnn_orig} & 83.11 & 81.45 & 91.49 \\
		GRCNN \cite{grcnn} & 84.21 & 81.09 & 92.28 \\
		Rosetta \cite{rosetta} & 84.08 & 81.94 & 92.15 \\
		RARE \cite{rare_paper} & 85.63 & 83.59 & 92.74 \\
		STAR-Net \cite{starnet} & 87.05 & 84.27 & 93.59 \\
		TRBA \cite{clovaai} & 88.92 & 85.61 & 94.16 \\ 
		\midrule[0.5pt]
		\multicolumn{4}{c}{\textbf{B.} Baseline OCR models (Transformer-based)} \\
		\midrule[0.5pt]
		Parseq \cite{parseq} & 26.13 & 25.60 & 26.41 \\
		ViTSTR \cite{vit_str} & 34.86 & 32.63 & 35.78 \\
		TrOCR \cite{tr_ocr} & 38.43 & 36.10 & 37.61 \\
		ABINet \cite{abi_net} & 41.17 & 40.20 & 38.96 \\
		CDistNet \cite{cdistnet} & 33.72 & 34.96 & 32.48 \\
		VisionLAN \cite{vision_lan} & 28.40 & 27.82 & 29.07 \\
		\midrule[0.5pt]
		\multicolumn{4}{c}{\textbf{C.} SOTA Urdu OCR models} \\
		\midrule[0.5pt]
            5LayerCNN-DBiLSTM \cite{arabic_hybrid_cnn_rnn_blstm} & 82.92 & 81.15 & 90.67 \\
		VGG-BiLSTM \cite{iiith17urdu} & 83.11 & 81.45 & 91.49 \\
		VGG-MDLSTM \cite{mdlstm_urdu_ocr,mdlstm_urdu_ocr2} & 83.30 & 81.72 & 91.17 \\
            VGG-LSTM-Attn \cite{arabic_attention_encoder_decoder_2021} & 84.16 & 82.21 & 91.88 \\
            VGG-DBiLSTM-Attn \cite{Arabic_attention_cnn_rnn} & 84.58 & 82.72 & 92.01 \\
		ResNet-BiLSTM \cite{resnet_urdu_ocr} & 86.96 & 84.18 & 93.61 \\
		DenseNet-GRU-Attn \cite{gru_attn_paper} & 91.10 & 85.32 & 94.63 \\
		\midrule[0.5pt]
		\ModelName-Small & 90.87 & 86.35 & 95.08 \\
		\ModelName & \textbf{92.97} & \textbf{88.01} & \textbf{95.97} \\
		\bottomrule[1.5pt] 
	\end{tabular}
	\vspace{1em}
	\caption{Performance comparison of SOTA OCR models}
	\label{tab:sota-comparisions}
	\vspace{-2em}
\end{table}

\begin{table}[b]
	\centering
	\setlength{\tabcolsep}{8pt}
	\begin{tabular}{lcccc}
		\toprule[1.5pt]
		\textbf{Model} & \textbf{Training Data} & \textbf{\DatasetNameReal} & \textbf{IIITH} & \textbf{UPTI} \\ 
		\midrule[0.5pt]
		\ModelName-Small & UPTI & 54.84 & 72.82 & 98.63 \\
		\ModelName-Small & \DatasetNameReal & 90.87 & 86.35 & 95.08 \\
		\ModelName-Small & \DatasetNameSynth &  75.14 & 85.47 & 92.09 \\
		\ModelName-Small & Mix-All & 91.71 & 90.04 & 98.72 \\ [0.5ex]
		\midrule[0.5pt]
		\ModelName & UPTI &  64.73 & 76.15 & \textbf{99.17} \\
		\ModelName & \DatasetNameReal &  92.97 & 88.01 & 95.97 \\
		\ModelName & \DatasetNameSynth &  80.26 & 87.85 & 93.49 \\
		\ModelName & Mix-All &  \textbf{93.39} & \textbf{90.91} & 98.36 \\ [0.5ex]
		\toprule[1.5pt] 
	\end{tabular}
	\vspace{0.5em}
	\caption{Performance after training with different datasets. See that training with UPTI dataset leads to poor performance on real-world validation datasets. (Here, ``Mix-All'' means a mixture of UPTI, \DatasetNameReal \& \DatasetNameSynth)}
	\label{tab:results_on_datasets}
	\vspace{-2em}
\end{table}

\begin{figure}[t]
	\centering
	\begin{tabular}{|c|c|}
		\hline
		\multicolumn{2}{|c|}{\bfseries Stage 1: Highest resolution feature maps} \\
		\hline
		\includegraphics[width=0.45\linewidth]{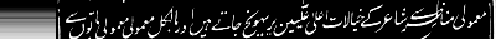} & \includegraphics[width=0.45\linewidth]{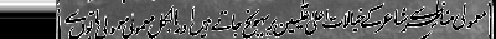} \\
		\includegraphics[width=0.45\linewidth]{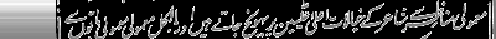} & \includegraphics[width=0.45\linewidth]{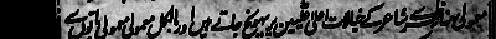} \\
		\includegraphics[width=0.45\linewidth]{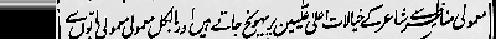} & \includegraphics[width=0.45\linewidth]{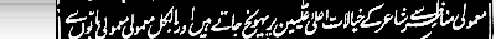} \\
		\hline
		\multicolumn{2}{|c|}{\bfseries Stage 4: Lowest resolution feature maps} \\
		\hline
		\includegraphics[width=0.45\linewidth]{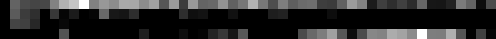} & \includegraphics[width=0.45\linewidth]{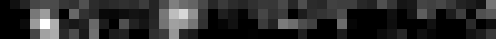} \\
		\includegraphics[width=0.45\linewidth]{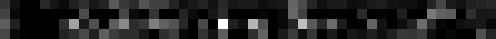} & \includegraphics[width=0.45\linewidth]{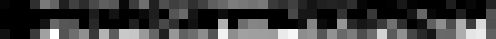} \\
		\includegraphics[width=0.45\linewidth]{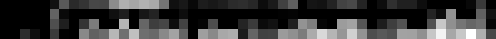} & \includegraphics[width=0.45\linewidth]{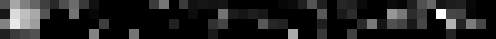} \\
		\hline
		\multicolumn{2}{|c|}{\bfseries Feature maps after concatenating all resolutions of Stage 4} \\
		\hline
		\includegraphics[width=0.45\linewidth]{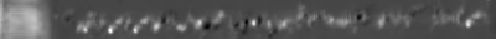} & \includegraphics[width=0.45\linewidth]{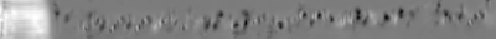} \\
		\includegraphics[width=0.45\linewidth]{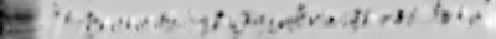} & \includegraphics[width=0.45\linewidth]{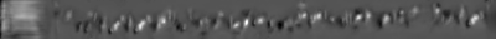} \\
		\includegraphics[width=0.45\linewidth]{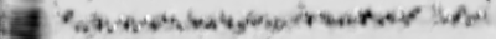} & \includegraphics[width=0.45\linewidth]{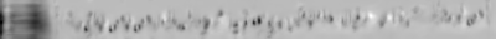} \\
		\hline 
	\end{tabular}
	\vspace{0.5em}
	\caption{Visualization of feature maps learnt by various layers of the \ModelName-Large. 
		The small features associated with various characters are lost at low resolution (middle row) but are present in the last row. This illustrates the effectiveness of our proposed high-resolution multi-scale feature extraction technique, leading to improved performance in the printed Urdu text recognition.}
	\label{tab:feature_vis}
\end{figure}

\section{Experiments And Results}
\label{section:experiments}


\mypara{Experimental Setup}
In order to ensure a fair comparison among the existing models in the field, we have established a consistent experimental setup for evaluating the performance of all the models and report all the results in Table \ref{tab:sota-comparisions}. Specifically, we have fixed the choice of training to the \DatasetNameReal training set, the validation set to be the validation sets of the datasets outlined in Figure \ref{fig:datasets_samples_and_stats}. Further, to compare the different available training datasets, we train our proposed \ModelName models on each of them and present the results in Table \ref{tab:results_on_datasets}. We have used the AdaDelta optimizer \cite{adadelta_orig}, with a decay rate of 0.95, to train our \ModelName models on an NVIDIA A100 40GB GPU. The batch size and learning rate used were 32 and 1.0, respectively. Gradient clipping was employed at a magnitude of 5, and all parameters were initialized using He's method \cite{hrnet_orig}. To improve the robustness of the model, we employed a variety of data augmentation techniques during the training process, such as random resizing, stretching/compressing, rotation, and translation, various types of noise, random border crop, contrast stretching, and various image processing techniques, to simulate different types of imaging conditions and improve the model's ability to generalize to real-world scenarios. The \ModelName-Large model achieved convergence in approximately 7 hours. We utilize the standard character-wise accuracy metric for comparison, which uses the edit distance between the predicted output (Pred) and the ground truth (GT):
\[
\text{Accuracy} = \frac{\sum \left( \text{length}(GT) - \text{EditDistance}(Pred,GT) \right) }{\sum \left( \text{length}(GT) \right)}
\]

\subsection{Results And Analysis}

In order to evaluate the effectiveness of our proposed architecture, we conducted a series of experiments and compared our results with state-of-the-art (SOTA) models for Urdu OCR, as well as a few for Arabic, including both printed and handwritten ones (Table \ref{tab:sota-comparisions}C). Additionally, we evaluated our model against the current SOTA baseline OCR models, primarily developed for Latin-based languages (Tables \ref{tab:sota-comparisions}A and \ref{tab:sota-comparisions}B). Our proposed model achieves superior performance, surpassing all of the SOTA OCR models in terms of character-wise accuracy on all three validation datasets, achieving a recognition accuracy of 92.97\% on the \DatasetNameReal validation set. It is worth noting that while Table \ref{tab:sota-comparisions}A presents a comparison of our proposed model against hybrid CNN-RNN models, Table \ref{tab:sota-comparisions}B presents a comparison against recent transformer-based models. The results clearly show that transformer-based models perform poorly in comparison to our proposed model and even the SOTA CNN-RNN models for Latin OCR. This can be attributed to the fact that these models, which are designed to be trained on massive datasets when applied to the case of Urdu script recognition, overfit the small-size training data and struggle to generalize, thus resulting in poor validation accuracy.

In our analysis of our proposed \DatasetNameReal and \DatasetNameSynth datasets against the existing UPTI dataset, which is currently the only available training dataset for this purpose, we found that our proposed datasets effectively improve the performance of the \ModelName model. When trained on the UPTI dataset, both \ModelName-Small and \ModelName achieve high accuracy on the UPTI validation set but perform poorly on the UTRSet-Real and IIITH validation sets. This suggests that the UPTI dataset is not representative of real-world scenarios and does not adequately capture the complexity and diversity of printed Urdu text. Our proposed datasets, on the other hand, are specifically designed to address these issues and provide a more comprehensive and realistic representation of the task at hand, as both of them perform significantly well on all datasets, with \DatasetNameReal being the best. Furthermore, the results show that combining all three training datasets can further improve the performance, especially on the IIITH and UPTI validation sets.

\begin{figure}[t]
	\centering
	\includegraphics[width=\linewidth]{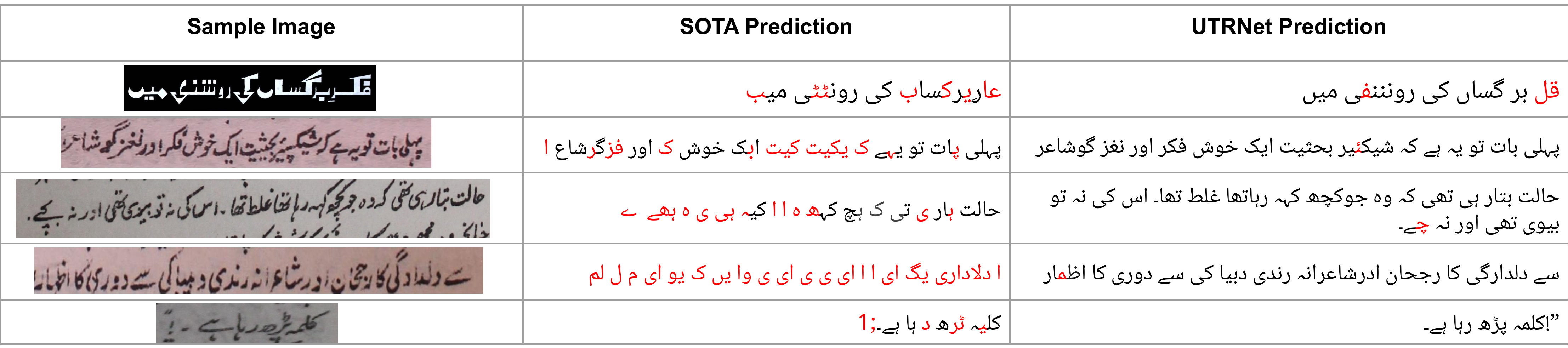}
	\caption{The figure illustrates a qualitative analysis of our proposed \ModelName-Small and the SOTA method \cite{iiith17urdu} on the \DatasetNameReal dataset. To facilitate a fair comparison, the errors in the transcriptions are highlighted in red. The results showcase the superior accuracy of \ModelName in capturing fine-grained details and accurately transcribing Urdu text in real-world documents.}
	\label{fig:qualitative_results}
\end{figure}

One of the key insights from our results is the significant difference in accuracy when comparing our proposed \ModelName model with the current state-of-the-art (SOTA) model for printed Urdu OCR \cite{iiith17urdu}, as presented in Figure~\ref{fig:char_accuracy}. This highlights the complexity of recognizing the intricate features of Urdu characters and the efficacy of our proposed \ModelName model in addressing these challenges. Our results align with our hypothesis that high-resolution multi-scale feature maps are essential for capturing the nuanced details required for accurate Urdu OCR. To further support this claim, we also present a visualization of feature maps generated from our CNN (as depicted in Figure \ref{tab:feature_vis}), which clearly demonstrates the ability of \ModelName to effectively extract and preserve the high-resolution details of the input image. Additionally, we provide a qualitative analysis of the results by comparing our model with the SOTA \cite{iiith17urdu} in Figure \ref{fig:qualitative_results}.

We also conducted a series of ablation studies to investigate the impact of various components of our proposed \ModelName model on performance. The results of this study, as shown in Table \ref{tab:ablation}, indicate that each component of our model makes a significant contribution to the overall performance. Specifically, since we observed that incorporating a multi-scale high-resolution feature extraction significantly improves the result for Urdu OCR, we tried various other multi-scale CNNs as a part of our ablation study. We also found that the use of generalisation techniques, such as temporal dropout and augmentation, further improved the robustness of our model, making it able to effectively handle a wide range of challenges which are commonly encountered in real-world Urdu OCR scenarios.

\section{Web Tool for End-to-End Urdu OCR}
\label{section:webtool}

\begin{wrapfigure}{r}{0.6\linewidth}
	\setlength{\columnsep}{0pt}%
	\centering
	\vspace{-2.5em}
	\includegraphics[width=1.0\linewidth]{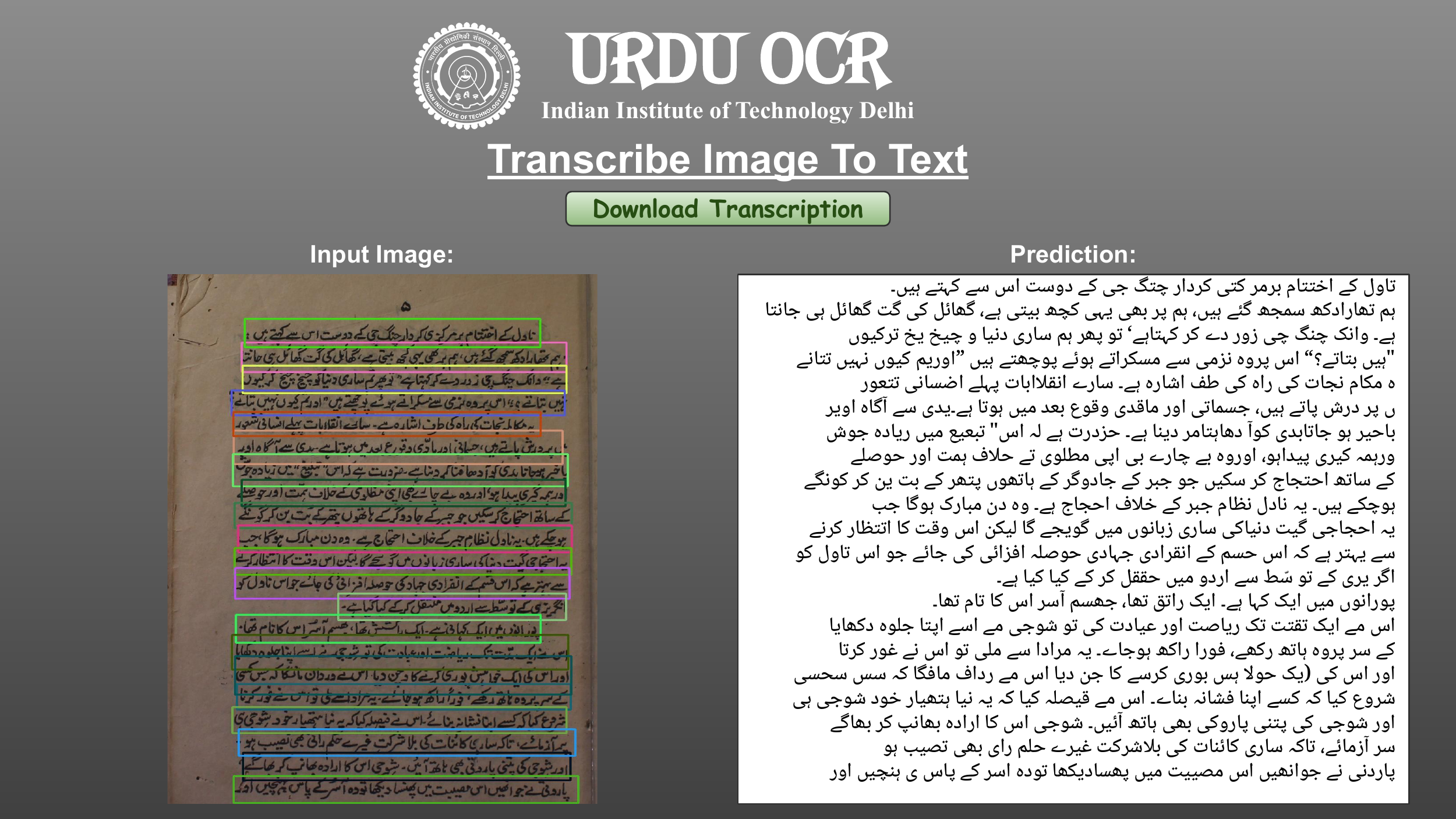}
	\vspace{-1.8em}
	\caption{Web tool developed by us for end-to-end Urdu OCR}
\end{wrapfigure}
We have developed an online website for Urdu OCR that integrates ContourNet \cite{contournet_orig} model, trained on the {\DetectionDataset} dataset with our proposed \ModelName model. This integration allows for end-to-end text recognition in real-world documents, making the website a valuable tool for easy and efficient digitization of a large corpus of available Urdu literature. 

%

\section{Conclusion}
\label{section:conclusion}

In this work, we have addressed the limitations of previous works in Urdu OCR, which struggle to generalize to the intricacies of the Urdu script and the lack of large annotated real-world data. We have presented a novel approach through the introduction of a high-resolution, multi-scale semantic feature extraction-based model which outperforms previous SOTA models for Urdu OCR, as well as Latin OCR, by a significant margin. We have also introduced three comprehensive datasets: \DatasetNameReal, \DatasetNameSynth, and {\DetectionDataset}, which are significant contributions towards advancing research in printed Urdu text recognition. Additionally, the corrections made to the ground truth of the existing IIITH dataset have made it a more reliable resource for future research. Furthermore, we've also developed a web based tool for end-to-end Urdu OCR which we hope will help in digitizing the large corpus of available Urdu literature. Despite the promising results of our proposed approach, there remains scope for further optimization and advancements in the field of Urdu OCR. A crucial area of focus is harnessing the power of transformer-based models along with large amounts of synthetic data by enhancing the robustness and realism of synthetic data and potentially achieving even greater performance gains. Our work has laid the foundation for continued progress in this field, and we hope it will inspire new and innovative approaches for printed Urdu text recognition.

\section{Acknowledgement}
\label{section:acknowledgement}

We would like to express our gratitude to the Rekhta Foundation and Arjumand Ara for providing us with scanned images, as well as Noor Fatima and Mohammad Usman for their valuable annotations of the \DatasetNameReal dataset. Furthermore, we acknowledge the support of a grant from IRD, IIT Delhi, and MEITY, Government of India, through the NLTM-Bhashini project.

\bibliographystyle{splncs04}
\bibliography{refs}

\end{document}